# KNOWLEDGE MANAGEMENT IN ECONOMIC INTELLIGENCE WITH REASONING ON TEMPORAL ATTRIBUTES

**Bolanle Oladejo** (*), **Adenike Osofisan** (**), **Victor Odumuyiwa** (*)
oladejof@loria.fr, nikeosofisan@gmail.com, victor.odumuyiwa@loria.fr

(*)LORIA, Campus Scientifique, Vandoeuvre les Nancy Cedex Nancy, France
(**)Department of Computer Science, University of Ibadan, Ibadan, Nigeria

**Mots clefs :**
Représentation de connaissances, gestion de connaissance, intelligence économique, raisonnement temporal, exploitation de connaissance, propriété temporal

**Palabras clave:**
Representación de conocimiento, administración del conocimiento, inteligencia colectiva, características temporales, razonamiento temporal, explotación del conocimiento

**Abstract**
People have to make important decisions within a time frame. Hence, it is imperative to employ means or strategy to aid effective decision making. Consequently, Economic Intelligence (EI) has emerged as a field to aid strategic and timely decision making in an organization. In the course of attaining this goal: it is indispensable to be more optimistic towards provision for conservation of intellectual resource invested into the process of decision making. This intellectual resource is nothing else but the knowledge of the actors as well as that of the various processes for effecting decision making. Knowledge has been recognized as a strategic economic resource for enhancing productivity and a key for innovation in any organization or community. Thus, its adequate management with cognizance of its temporal properties is highly indispensable. Temporal properties of knowledge refer to the date and time (known as timestamp) such knowledge is created as well as the duration or interval between related knowledge. We observe that a number of existing systems are concerned much about knowledge resources with less emphasis on its temporal properties. This paper focuses on the needs for a user-centered knowledge management approach as well as exploitation of associated temporal properties. Our perspective of knowledge is with respect to decision-problems projects in EI. The central aim of EI is to obtain tactical and strategic knowledge in order to facilitate timely, cost-effective decision-making process. Thus, the acquisition, representation and storage of knowledge with its temporal properties are required to exploit it for future use. Our hypothesis is that the possibility of reasoning about temporal properties in exploitation of knowledge in EI projects should foster timely decision making through generation of useful inferences from available and reusable knowledge for a new project.

# 1. Introduction

Virtually all events in life are time-stamped. In other words, there is time associated with every situation be it task, process, event etc. As essential as knowledge is so is the temporal properties in any given context. The problem of representing knowledge and temporal reasoning arise in a wide range of disciplines, including Computer science, Medical Science, Philosophy, Psychology and Linguistics. In the field of Artificial Intelligence (AI), a branch of Computer Science: temporal properties (time point, interval, etc.) on facts, events and processes are important for certain activities like planning, prediction and management of such processes (McDermott, 1982; Allen, 1984). We believe that reasoning about temporal properties of knowledge in knowledge management system, in the context of Economic Intelligence (EI) would enhance the efficiency of such system. Also in the course of applying stored knowledge to a new problem case, such type of knowledge exploitation, can guide the plan of resolution of the new problem. In our work, we consider the two main categories of knowledge - explicit and tacit knowledge. Explicit knowledge comprises the documents, events, procedures etc. while actor's skill or know-how, understanding, observation etc. constitute tacit knowledge. These knowledge could be elicited for representation and modeling.

The two main concepts of this paper are economic intelligence and temporal reasoning.

EI can be defined, at the level of a particular enterprise, as the decision-maker's capacity to exploit the knowledge and the new experiences, while reinvesting those already acquired, in order to solve at best a new decision problem (Kislin, et al., 2002).

We observe that there are existing works as regards reasoning about temporal properties of knowledge. McDermott (1982) propounds reasoning based on time points of event or fact. The reasoning formalism in the work of Allen (1984) is based on the interval between events. Our focus differs from theirs due to the fact that we concentrate on knowledge management: that is, representing, organizing and storing of knowledge of a given project for future exploitation in the context of EI.

## 1.1 Structure of the paper

The paper is introduced in section one. A concise review of concepts of knowledge, Knowledge Management, Economic Intelligence and temporal reasoning including an evaluation of the features of existing systems that applies temporal reasoning as well as Knowledge Management systems are presented in section two. Our approach of reasoning on temporal properties in the exploitation of knowledge of projects in the context of Economic Intelligence is the focus of section three. Section four presents the architecture for representation of knowledge resource involved in decision-problem resolution in the context of EI. The paper is concluded in the last section with a summary.

# 2. Theoretical Background

We examine here the basic concepts that relate to knowledge, Knowledge Management, Economic Intelligence and Temporal Reasoning.

## 2.1 Knowledge

We need to distinguish knowledge from information and data in order to acquire a true perception of its significance in KM. From the point of view of Data Processing in Computer Science, data is generally referred to as raw facts or figures; e.g. Esther: name of a person, 10: age etc. Information on the other hand is the result of processed or related data; e.g. 'Esther is 10 years old'. According to Pohl (2002), it is defined in terms of data as numbers and words with relationships. Information is equated to knowledge in some literature. Knowledge is defined as the remembering of previously learned material or information; ranging from specific facts to complete theories (Bloom, 1956). In Artificial Intelligence, it is defined as related facts and situations of a domain in real world that could be represented. Diverse definitions of knowledge exist in related literature ranging from philosophical (epistemology) point of view to Computer science's (Artificial intelligence) perception. Pohl (2002) refers to knowledge as the addition of context to information. Considering the stated example about the age of a person, there is need to interpret what 'age' means before the information can be transformed into knowledge. For instance, we can depict age by the last birthday given the date of birth of a person. Thus, Esther's age can be a fact or knowledge when her date of birth is given and the current date is considered. From this analogy, we can define knowledge as facts with its attributed meaning, where meaning is a function of an observation, learning, experience, and understanding of a reality in a particular situation or context at a specific period of time by an individual. This view is illustrated in Figure 1. We consider next the classification of knowledge.

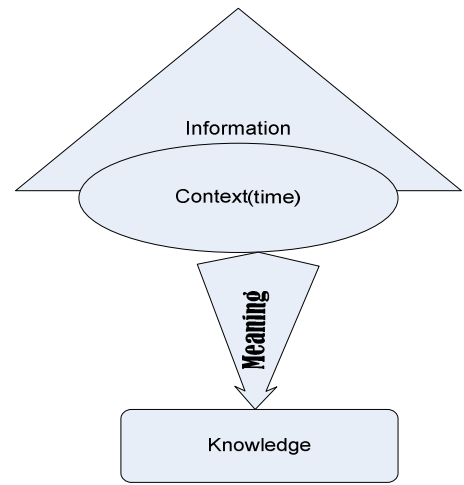

*Figure 1: what is knowledge?*

### 2.1.1 Types of Knowledge

Knowledge is classified into explicit or theoretical and tacit or practical knowledge (Nonaka & Takeuchi, 1995). Explicit knowledge could be expressed in form of theoretical and practical experience. It could be readily transmitted across individuals formally and systematically. On the other hand, tacit knowledge is in form of skill and it is highly personal and hard to formalize, thus, difficult to share with others. There is need to transform tacit knowledge into explicit knowledge in order to represent it in a way to aid its exploitation in a knowledge management system.

## 2.2 Knowledge Management

Knowledge Management (KM) is considered as a tool for competitive advantage and innovation for an organization, institution, industry or government. KM focuses on both theoretical and practical know-how of groups of people in an organization. There are diverse definitions of KM. It is defined as a continuous process of knowledge explicitation and internalization (Nonaka & Takeuchi, 1995). In addition, KM refers to *a global process in the enterprise, which includes all the processes that allow capitalization of knowledge capital of the firm. It focuses on the understanding, supporting, optimizing and accelerating those processes, in coherence and cross-fertilization in order to aid the prediction of future events from the knowledge of the past and present events, that is, "to know where you are and where you come from to better know where you go"* (Dieng-Kuntz & Matta, 2002). Succinctly, KM implies *the explicit and systematic management of vital knowledge - and its associated processes of creation, organization, diffusion, use and exploitation* (Skyrme, 1998). We refer to KM as the process of conscious co-ordination of knowledge - skills, expertise, procedures, etc.; in order to store, organize and exploit it for innovation and optimization of organizational goals. Consequently, the major requirements of KM are creation, acquisition, organization, storage and exploitation of knowledge. We consider the effective principles of knowledge representation and capitalization with respect to the highlighted criteria below.

- ➢ Co-operative knowledge acquisition or elicitation between KM system developer and domain actors.
- ➢ User-centered representation format for presentation of system result to users.
- ➢ Adequate knowledge retrieval method which supports ease of system use and access to relevant and reliable knowledge by users.
- ➢ System flexibility to support evolution of KM system.
- ➢ Adequate method of evaluation or assessment of a knowledge repository in order to improve its efficiency and adequacy.

KM usually involves the development of knowledge repository generally called corporate or organizational memory. Corporate memory (CM) refers to a structured set of knowledge related to the firm experience in a given domain (Simon, 1996). It is also a "repository of knowledge and know-how of a set of individuals working in particular firm" (Euzenat, 1996). In building a CM, it is essential to identify the crucial expert or domain knowledge in order to determine the required kind of CM that would "support the integration of resources and know-how in the enterprise and the co-operation by effective communication and active documentation" (Durstewiz, 1994). Some examples of methodologies for developing CM are Case-Based Reasoning (CBR) approach, Ontology, corporate Semantic Webs, agent-based approach etc. CM can be in form of knowledge-based, document-based, workflow-based, and distributed Organizational Memory systems (Dieng, et al., 1998). KM is applicable to EI since it serves as a mean of organization, conservation, and transmission of existing knowledge in EI projects. There are several application areas where KM has been implemented as an innovation key. We briefly consider some of these systems.

### 2.2.1 Application of KM

A number of applications already exist in the field of knowledge management or capitalization. An example of such applications is the capitalization of the knowledge of how to synthesize "purely Swiss" vitamin C (Bachi, 2005): with emphasis on its impact on society through the influence of technology. There is another system on capitalization of design process (Matta et al., 2002); it focused on the development of design project memory from designers' activities through direct extraction and tracking of knowledge from project design tools, design process and product data. Other applications are capitalization of business experience and resources (Gaynard, 2004), of steel production process and defect (Simon, 1996), of industrial systems (Admane, 2005), of equipment diagnosis and repair help system (Chebel-Morello et al., 2005). In all these work and a host of others, there is no emphasis on the issue of temporal properties of events and knowledge that is capitalized. Hence we address this issue in the context of KM of EI projects.

## 2.3 Economic Intelligence

Economic Intelligence is a set of coordinated actions of search, processing and distribution for exploitation, of useful information for economic actors. These actions are carried out legally with all the necessary protection of the company's patrimony, and with the best quality, delay and cost (Matre, 1994). Also, it is the process of collection, processing and distribution of information with the goal of reducing uncertainty in taking strategic decisions (Revelli, 1998). Thus, it concerns all coordinated actions for research, treatment and distribution, for its operation, useful information to economic actors; carried out legally with all the guarantees of protection necessary for preserving the intellectual heritage of the company, in the best conditions for quality, time and cost. It involves the whole process by which a decision-maker has a good understanding of the land on which he operates on the basis of available resources to make policy decisions (Carayon, 2003). There are basic steps required in achieving this goal.

The EI research develops methods for identifying relevant sources of information, analyzing the collected information and manipulating it to provide what the user needs for decision making. There are three major economic actors involved in the process:

- ❖ **decision maker** who must formulate exact description of the decision-problem;
- ❖ **watcher** (person charged with obtaining relevant problem associated with the decision problem at hand) who must locate, supervise, validate and emphasize the strategic information needed for solving the problem, and
- ❖ **project coordinator** who serves as a link between decision-maker and watcher as well as end-user (Knauf, 2007).

All the actors work together in collaboration (Odumuyiwa & David 2008) in order to optimize sharing of strategic knowledge amongst one another. Thus, EI is decision-maker-centered as information needs are treated based on the contexts of the decision problem at hand. There is need to identify and acquire for storage all knowledge resources involved in the EI projects and the actors' know-how. Also, tacit knowledge or know-how of actors need be converted into explicit knowledge before it can be represented for storage. There are existing models (SITE activity report, 2007) for representing the actors' knowledge and their activities. They aid the conversion of tacit to explicit knowledge. These models are to be integrated into a single model for capturing and storage of knowledge in a given decision making project. We attempt to apply KM principles to accomplish this requirement for capitalizing identified knowledge and its temporal properties to foster subsequent decision making. First of all, we identify knowledge resources in EI.

The process of EI is a systematic approach which harnesses seven stages to resolve decision-making problem. According to (Bouaka & David, 2004) it entails the phases highlighted below (figure 2).

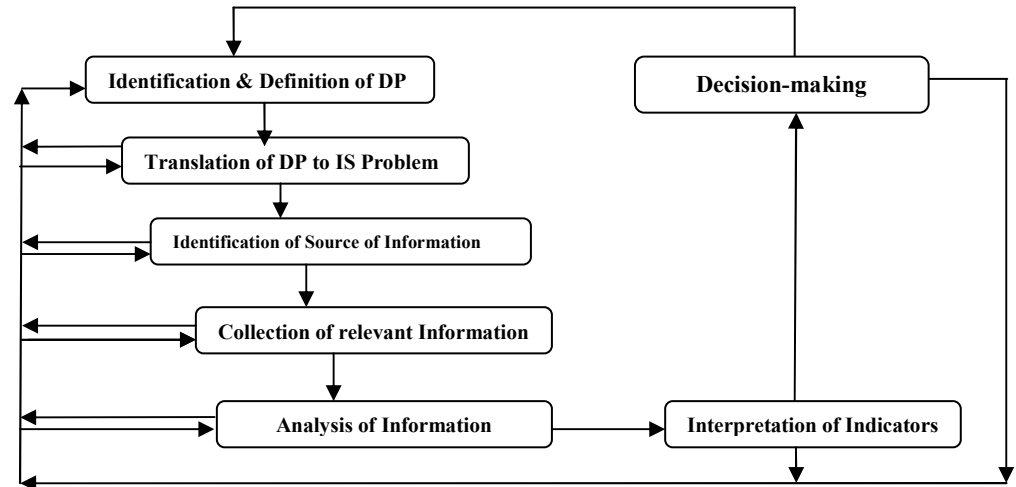

*Figure 2: EI process*

a) Identification and definition of a decisional problem (DP): This phase involves analyzing a decision problem by examining the meaning, purpose and the context of the problem in order to get the actual needs of a decision-maker.
b) Translation of the decisional problem to information search (IS) problem: Definition of a decisional problem is succeeded by formulation of related information research questions or problems that will pilot the resolution process of the identified decision problem.
c) Identification of relevant information sources: This phase enables goal-oriented and user-centered search for desired information from both internal and external environment of a decision problem.
d) Collection of relevant information: Information from sources such as, databases, documents, experiment results, environmental scanning result etc. are gathered for validation.
e) Analysis of the collected information to extract indicators for decision: This phase requires thorough examination or assessment of collected information in order to generate value-added information through annotation.
f) Interpretation of indicators: All the economic actors are involved in this phase. They verify the validated information for suitability of resolving the decision problem.
g) Decision-making: The concession of actors culminates to precise resolution of an identified problem.

It is possible to retrace each phase to any of the preceding phase(s) for validation purpose.

Below are some models proposed by the research team SITE[1] at LORIA[2] in Nancy France for representing the knowledge of the phases in EI process.

- ❖ Model for Explicit Definition of Decision Problem (Bouaka, 2004).
- ❖ Model for Watcher's Information Search Problem (Kislin, 2007).
- ❖ Model for Information Retrieval query Annotations Based on Expression (Goria, 2006).
- ❖ Model for Representation of Information Search Problem in Economic Intelligence (Afolabi 2007).
- ❖ Annotation Model for Information Exchange (Roberts, 2007).
- ❖ Model for Identification and Representation of User's needs at the time of Interrogation of Information system (Peguiron, 2006).
- ❖ Model for Specification of the competence of the Coordinator of Regional EI (Knauf, 2007).
- ❖ Communication model for Collaborative Information Retrieval (Odumuyiwa & David, 2008).

These models are proposed independently. In order to exploit the experience and results of past decision-problem resolution project(s) it is necessary to integrate and represent these models. One of the approaches for exploiting knowledge in EI is to consider the use of temporal reasoning.

## 2.4 Temporal Reasoning

Temporal reasoning is vast and dominant in diverse AI applications, especially in the areas where time-stamped events are modeled. Fundamentally, reasoning on temporal properties was initiated with the paradigm of time point of facts and events propounded by McDermott (1982) for reasoning about continuous change, processes, causality and for planning actions. McDermott's dichotomy refers to *facts* and *events*, as the two basic entities that are associated with time. A fact is a set of states, intuitively those in which it is true. An infinite collection of states (or points) is introduced as the set of primitive temporal elements. Each state $s$ has a time of occurrence, $d(s)$, a real number called its date, where states are partially ordered by the "*no later than*" relation $\leq$. The time structure is characterized as linear from the past and branching into the future, and useful for modeling possible worlds in action planning. In order to model continuous change, it is assumed that between any two distinct states, there is a continuum of states. This is achieved by modeling each single branch, called chronicle, as a connected series of states, which is isomorphic to the real line. He uses formula (T $s$ $p$) to indicate that fact $p$ is true in state $s$, while the notion of $p$ being true over the time interval between states $s_1$ and $s_2$ is written as (TT $s_1$ $s_2$ $p$); where T is temporal predicate and p refers to a given proposition or statement.

---

[1] SITE (Modeling and Development of Economic Intelligence Systems)

[2] LORIA (Laboratoire Lorraine de Recherche en Informatique et ses Applications)

Reasoning about time intervals of processes and events is propounded by Allen. His theory of time and action is based on intervals as primitive rather than as derived structure from points (Allen, 1984) as in McDermott's case. A set of 13 mutually exclusive binary relations between two intervals are introduced, i.e., *EQUALS, BEFORE, AFTER, MEETS, MET_BY, OVERLAPS, OVERLAPPED_BY, STARTS, STARTED_BY, DURING, CONTAINS, FINISHES and FINISHED_BY*. Unlike McDermott's dichotomy of facts and events with respect to time, Allen introduces three ontological categories, i.e., *properties*, *events* and *processes*, to time intervals over which they hold or occur. He denotes the assertion that property $p$ holds over interval $i$ by the formula HOLDS $(p, i)$, using axiomatic theorems.

Ma and Knight (1996) proposed Reification approach which serves as a sort of bridge with respect to synthesis of both McDermott's and Allen's theories; as well as to semantics (power of expressiveness) of logical assertions. It allows one to reason about the truth of assertions over time while preserving the first-order structure of the propositions. It is meant for representing and reasoning about temporal and non-temporal relationships between non-temporal assertions, and for expressing the temporal relations between actions and effects. We consider next a number of applications that adopt temporal reasoning.

### 2.4.1 Existing Systems applying Temporal Reasoning

Chen worked on uncertain temporal knowledge management (Chen, 1992). The emphasis is on the use of fuzzy logic to handle uncertain (imprecise time-stamp) temporal events: represented in form of possibility distributions. KM involves the creation, acquisition, organization, storage and exploitation of knowledge. The author claimed to manage temporal knowledge but his approach is different from that of KM practice since his focus was mainly on computing intervals between imprecise periods (several hours, long time etc.) of processes' events.

Another direction is that which centers on reasoning about processes in distributed system with emphasis on the sets of states of processes, runs (execution of network process or protocol) and time points of the runs (Fagin & Halpern, 1994). Their approach is based on Kripke structure (Kripke, 1963) which formalizes intuitions behind event occurrence (possible world). They were able to model and prove the correctness of the probability spaces among processes' states using a language of axiomatization. Also, decision procedures are obtained from the reasoning process. The common feature of their work with this one is the fact that both reason on knowledge i.e. events and also consider time points of events' states. However, the discrepancy is that there is no provision for capitalization of knowledge. An apparently similar application is another instance of workflow system.

*Workflow system is a complete or partial automation of a business process, in which participants (humans or machine) are involved in a set of activities according to certain procedural rules and constraints* (Bettini et al., 2002). Their work aimed at reasoning about the quantitative temporal constraints on the duration of activities and the required synchronization in order to enhance capability of workflow systems. However, they only concentrate on the temporal issue of workflow specification. The notion and framework of flexible temporal properties is applied to scheduling of activities in an office environment and for representing preferences among solutions and possible priorities of the temporal properties (Badaloni et al., 2000). Another system which integrates temporal reasoning and maintenance in medical applications is considered next.

Shahar (1999) demonstrated the importance of managing (storage, retrieval and reuse of) time-oriented clinical data by presenting an architecture called a *temporal mediator*. It combines temporal reasoning and temporal maintenance by integrating the management of clinical databases and medical knowledge bases. In conclusion, all these systems mainly concentrate on the temporal issue with less attention to management of knowledge.

Temporal properties such as date and time of facts enable the possibility of making useful inference or assertion from such facts which could further aid or guide certain decision related to the context of the facts. However, if temporal attributes should be treated in isolation of knowledge it cannot in any way accomplish the goal of KM which bores on reuse of knowledge for a new problem. We discover that the knowledge resources in the context of EI could be assigned respective date-time stamps. Consequently, exploitation of the knowledge will include generation of useful inferences from available and reusable knowledge for a new project by reasoning about associated temporal properties. This functionality in exploitation of knowledge in projects of decision-problem resolution would foster adequate planning with respect to schedule of application of knowledge to a new problem.

## 3. Knowledge Management in EI with a view of reasoning on temporal properties.

EI is an indispensable decision-support system in which problem resolution are driven by users' needs. KM and EI are complementary and we can rightly attribute them as a couple. KM serves as a tool for enhancing productivity and innovation in an organization; so does EI likewise fosters competitive advantage resulting from effective, timely and economical decision-making process. The goal of both EI and KM is provision of adequate knowledge to appropriate users.

We have adopted the approach by which knowledge is specified with date/time stamps attribute. This serves as an important fragment of knowledge as we discover that it aids reasoning about the temporal properties of knowledge. Temporal Reasoning (TR) actually enables planning specifically and decision-support task generally. Due to this inherent capability of TR, we consider a KM system for EI which basically captures and stores knowledge in decision problems resolution of an application or domain with the associated time stamps. In essence, it is possible to reason, not only about the date/time stamp of the knowledge base, but also about the interval between the elements of knowledge. We consider an example of a decision problem on 'Traffic go-slow' in a developing country. The goal of this problem is to identify the root cause and to recommend feasible solution by application of strategic methods of EI.

The knowledge in the resolution of the problem includes:
- ➢ Understanding the meaning of 'traffic go-slow' in the given context including the initial date of identification of problem and respective date of evolution of its interpretation by decision maker(s).
- ➢ Specification of the Stake of the problem and its date (stake signifies what to be gained if the problem is treated or lost if otherwise).
- ➢ Specification of appropriate information search need for resolving the decision problem including its date by watcher(s).
- ➢ Validation process on the specification of the information need and the date stamps of each interpretation task.

These knowledge resources are all contextual and time-stamped. That is, any of the knowledge occurs within a timeframe and as a result of certain factors, e.g. definition of traffic go-slow may span throughout the period of its interpretation. Invariably, it is possible to reason about the 'chronicle' or history of related knowledge with respect to their contexts and timestamps. In the above example, information on the problem, the process of resolution and the interval between each stage could be extracted. In order to achieve our proposal of reasoning using temporal properties in KM system of EI projects, we subsequently developed a framework which is generic and adaptable for decision-support system in any application generally, and specifically within the context of EI.

## 4. Framework for KM of EI

Figure 2 depicts a framework for the representation of knowledge involved in decision-problem resolution in the context of EI. Our approach is to model KM from user's perspective in order to address: "*Who*" does "*What*" "*When*", "*Why*" and "*How*". The framework permits the reasoning using the temporal properties (denoted by t for date/timestamp and the subscripts i, l, f, for initial, later and final respectively) of knowledge.

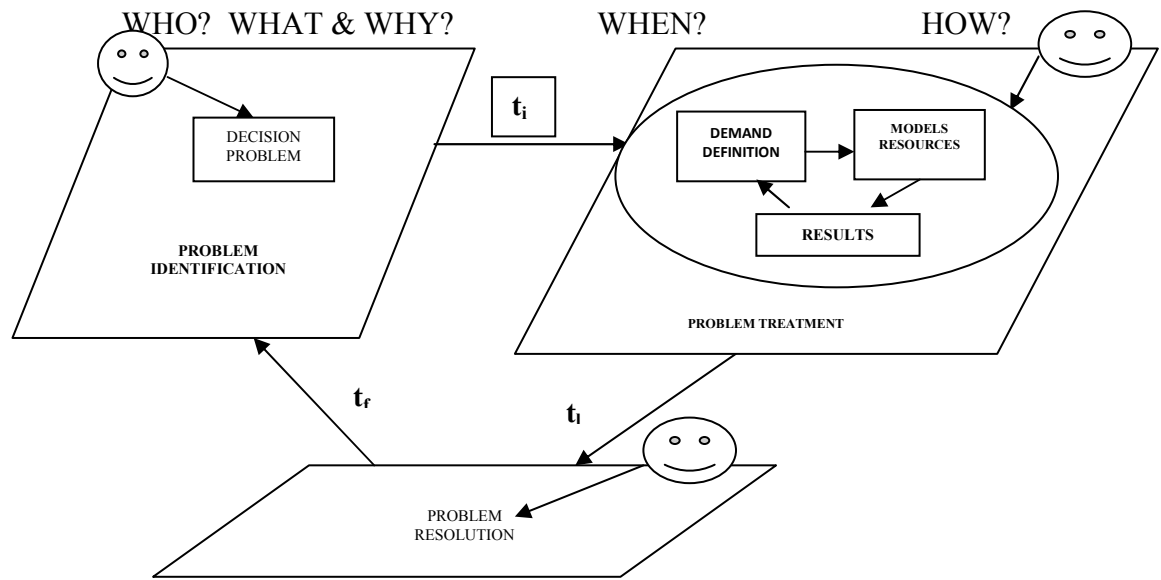

*Figure 3: Framework of EI knowledge repository with temporal properties.*

In figure 2 above, the framework presents a platform for representing knowledge with respective date/timestamp of EI decision-problem projects. Considering the decision problem given in section 3, the decision maker(s) – *(representing who)* identifies the decision problem as traffic go-slow *(representing what)*. The reason for the decision problem depicted by *why* results from the effects of *what* i.e. slowing down of traffic in the given context at a particular time – i.e. *when*. The problem is addressed and resolved applying the phases of EI process (i.e. *how*) by designated actors at specific date/time. Finally, the decision problem is resolved by interpretation of indicators extracted from relevant information at a specific date/time or over a period. We implement the architecture in a simple prototype system.

## 4.1 Implementation of the framework in a prototype

We are able to build a knowledge repository based on the framework. We implement the system in an intranet environment in form of a knowledge portal. The exploitation of knowledge with temporal properties facilitates the retrieval of knowledge with their respective timestamps and reasoning on duration or time interval between related knowledge. The specification of the prototype system for realizing this goal is highlighted below.

a) Exploration of knowledge is based on the role of an actor.
b) User's activities are captured in terms of **who? what? why? and when?** The user is guided to indicate his need using a query method based on analysis and constraints (David & Thiery, 2003) in order to exploit stored knowledge that relates to a new problem.
c) Exploitation of knowledge with their respective timestamps.
d) Analysis of duration between stored knowledge.

The inclusion of temporal properties of knowledge aids qualitative reuse of knowledge in a KM system. For instance, exploitation of knowledge with their respective timestamps can help users or actors to access required information based on deduction from analysis of the timestamps. It could facilitate access to related decision problems in a specified range of period e.g. year. It could aid to determine the proficiency of an actor by granting access to the period such an actor has participated in resolving decision problem. Consequently, an actor can be guided with respect to the plan of application of re-usable knowledge to a new project. Moreover, we take cognizance of different contexts of users' needs which change considerably over time. This issue is taken care of in the KM system by handling exploitation of knowledge based on the roles of actors and contexts of need.

## 5. Conclusion

KM is inevitable for innovation in organization and EI also serves as an instrument for competitive advantage in terms of strategic decision making. Knowledge is an economic resource which needs be acquired from both activities and actors in a given project, for storage and exploitation. There is less emphasis on exploitation of temporal properties of knowledge in many KM systems. Consequently, we proposed to exploit temporal properties in knowledge management systems of EI projects in order to reason on available knowledge to generate useful inferences for a new project.
An important requirement of developing a dynamic KM system is the consideration of user needs. We adopt an approach in which the role of user or actor drives the system's results. EI aims to provide relevant information to right user(s) for solving decision problem. Thus, the major goal of

EI is likewise satisfied. The provision of relevant information and the possibility of making inferences based on time of occurrence associated with stored knowledge resource serve as an advantage for addressing future problems in decision making process of an organization.